\documentclass{article} 
\usepackage{iclr2027_conference,times}

\usepackage{amsmath,amsfonts,bm}

\def\eqref#1{equation~\ref{#1}}

\def\1{\bm{1}}

\DeclareMathAlphabet{\mathsfit}{\encodingdefault}{\sfdefault}{m}{sl}
\SetMathAlphabet{\mathsfit}{bold}{\encodingdefault}{\sfdefault}{bx}{n}

\usepackage{graphicx}
\usepackage{booktabs}
\usepackage{tabularx}
\usepackage{xcolor}
\usepackage{tikz}
\usetikzlibrary{arrows.meta,positioning,fit,backgrounds,calc}

\usepackage{hyperref}
\usepackage{url}

\title{Agentic High-Dimensional Bayesian Optimization with Hypothesis- and Evidence-Guided Search}

\author{
Zhixuan Gao\textsuperscript{1,2},
Ke Xue\textsuperscript{1,2}\thanks{Corresponding authors: \texttt{xuek@lamda.nju.edu.cn} and \texttt{qianc@lamda.nju.edu.cn}.},
Rongxi Tan\textsuperscript{1,2},
Ming Chen\textsuperscript{1,2}
\& Chao Qian\textsuperscript{1,2}\footnotemark[1] \\
\textsuperscript{1}State Key Laboratory of Novel Software Technology, Nanjing University\\
\textsuperscript{2}School of Artificial Intelligence, Nanjing University\\
\texttt{\{gaozx,xuek,tanrx,chenm,qianc\}@lamda.nju.edu.cn}
}

\definecolor{draftblue}{RGB}{35,85,130}

\newcommand{\method}{\textsc{HERA}}
\newcommand{\engine}{\textsc{PRISM}}

\iclrfinalcopy 
\begin{document}

\maketitle
\fancyhead[L]{Preprint. Under review.}

\begin{abstract}
High-dimensional Bayesian optimization (HDBO) seeks sample-efficient optimization when the number of variables is large relative to the evaluation budget. Recent LLM-based and agentic BO methods incorporate task knowledge and adapt search decisions during a run, but have primarily been evaluated on low- and moderate-dimensional problems. 
We ask whether this paradigm can transfer to the higher-dimensional regime. 
Our experiments show that these methods do not remain reliable in the high-dimensional regime, where the challenge is not only where to evaluate, but also which modeling assumption and search geometry to use when the objective's useful structure is unknown. 
We therefore introduce HERA, a Hypothesis- and Evidence-guided Research Agent that uses task context, optimization feedback, and structural diagnostics to revise search hypotheses, select and configure HDBO strategies, and determine their execution length. PRISM, its numerical optimization engine, generates and evaluates candidates sequentially within each search block, updating numerical models after each observation.
HERA remains competitive with strong numerical HDBO baselines and outperforms the evaluated LLM-based and agentic methods on four metadata-free synthetic functions.
Across eight real-world tasks, HERA achieves the best mean final objective among all evaluated systems on most benchmarks.
Further analyses show that structural diagnostics change strategy use, metadata effects vary across tasks, and adaptive search blocks reduce inference cost. 
\end{abstract}

\section{Introduction}
\label{sec:intro}

Bayesian optimization (BO) is a standard approach to optimizing expensive black-box functions under limited evaluation budgets~\citep{bosurvey1,bosurvey2}. Task descriptions and prior experimental experience can inform this search, but their implications are difficult to encode in numerical optimization rules. Recent LLM-based methods augment BO with task descriptions and other informal priors~\citep{llambo,labo}, while agentic BO places an LLM agent in control of an evolving optimization campaign and retains a Bayesian backend for uncertainty-aware search~\citep{sara}. This division combines the agent's ability to interpret heterogeneous context and revise decisions with the systematic numerical search that makes BO sample efficient.

These systems have been evaluated mainly on low- to moderate-dimensional problems, leaving their applicability to high-dimensional BO (HDBO) underexplored~\citep{opro,llambo,hollm,labo,sara}. In HDBO, sparse observations make surrogate modeling and candidate search difficult, and numerical methods introduce different biases---including coordinate sparsity, low-dimensional embeddings, additivity, and locality---to make search tractable~\citep{lassobo,alebo,addgpucb,turbo,hdbo-survey2}. Recent results also show that carefully designed full-dimensional BO can be competitive~\citep{vanilla-0,vanilla-understanding}, so no single strategy is an obvious default. We therefore ask whether agentic BO can transfer to high-dimensional problems, where limited evaluations must guide both objective optimization and the choice of search strategy.

This uncertainty creates two obstacles to a direct transfer of Agentic BO. First, the numerical backend itself is not uniformly reliable: a generic full-dimensional backend can become fragile, while a fixed specialized backend commits in advance to a bias that may be mismatched to the objective~\citep{vanilla-understanding,gitbo}. Second, delegating candidate generation to an LLM does not remove the difficulty; complete high-dimensional proposals can provide poor coverage or become effectively uninformative as coordinate-wise decisions accumulate~\citep{hollm}. These obstacles establish a transfer gap rather than a universal failure of Agentic BO, and raises two related questions: \emph{what should an LLM control in HDBO, and how should that control be exercised while retaining reliable numerical search?} Our answer is to place the agent above candidate generation: it manages revisable search hypotheses and bounded numerical experiments, while specialized methods decide which concrete points to evaluate.

\begin{figure}[t]
  \centering
  \includegraphics[width=1\textwidth]{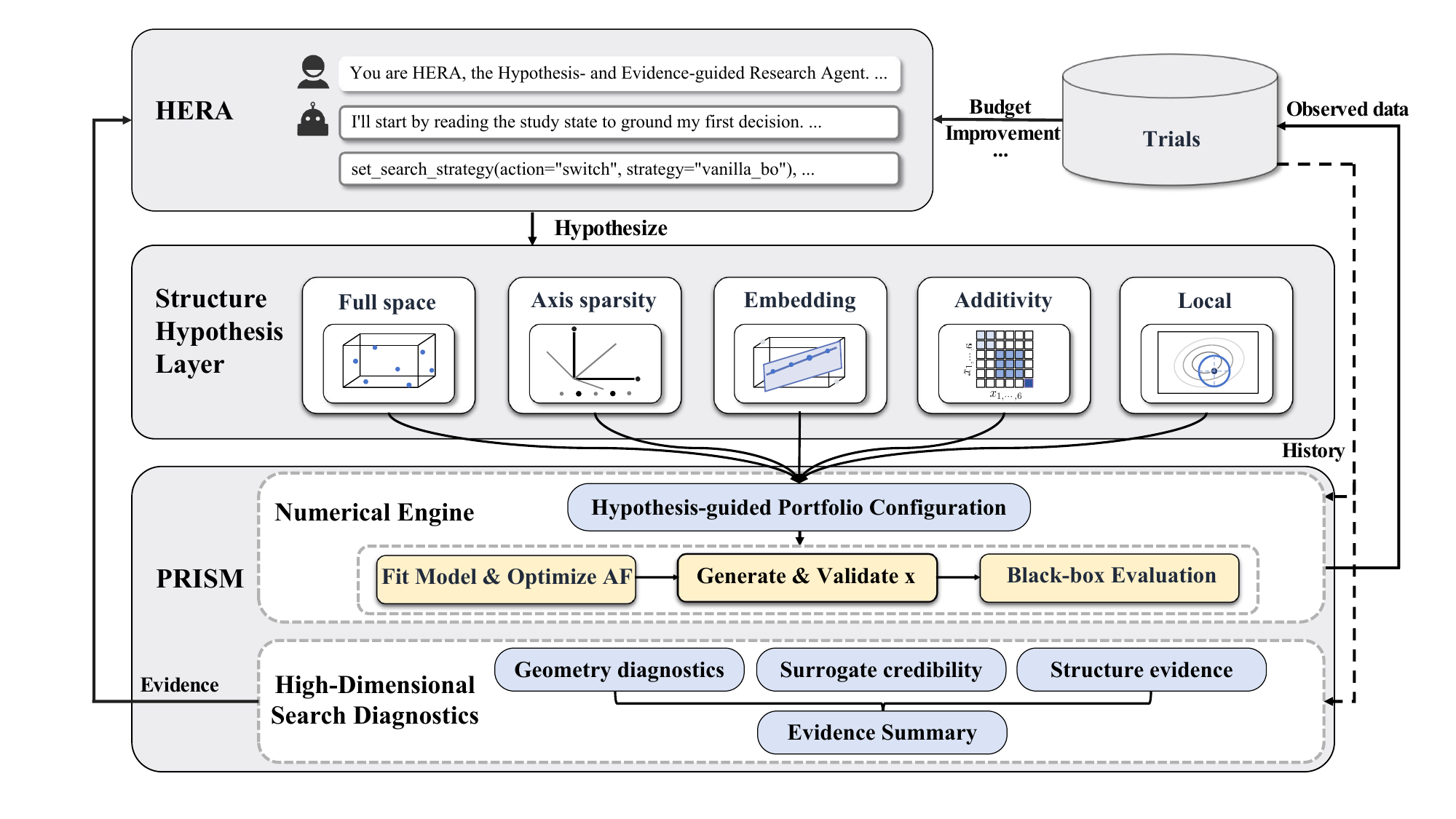}
  \vspace{-3em}
  \caption{\textbf{Agentic HDBO as an iterative research process.} \method{} interprets task information, optimization feedback, and structural diagnostics to maintain a revisable working hypothesis. \engine{} executes the selected strategy for several evaluations and updates a shared observation history. Outcomes return as evidence for the next round of hypothesis revision.}
  \vspace{-1.5em}
  \label{fig:method}
\end{figure}

We instantiate this idea in \method{}, a Hypothesis- and Evidence-guided Research Agent, and \engine{}, a Portfolio of Reconfigurable Inference and Search Methods (Figure~\ref{fig:method}). \method{} interprets task context, optimization progress, and provisional structural diagnostics to maintain a working hypothesis and choose whether to continue, reconfigure, restart, or change a numerical strategy. \engine{} validates that intervention, generates and evaluates candidates through the selected HDBO method, and retains all observations in a shared history. The resulting \emph{observe--hypothesize--intervene--evaluate--reflect} loop allows one agent decision to govern several numerical evaluations before reassessment. A working hypothesis is an actionable and revisable account of what may currently make the objective searchable, not a claim that the true structure has been recovered.

Experiments characterize our approach across four metadata-free synthetic functions and eight real-world tasks spanning trajectory planning, continuous control, engineering design, chip placement, and molecular optimization. Across synthetic and real-world benchmarks, \method{} matches strong classical high-dimensional BO methods, outperforms the evaluated LLM-based and agentic baselines, and achieves the best mean final objective among all evaluated systems on the majority of real-world tasks. We further examine whether simpler strategy selection is sufficient, and ablate the roles of structural diagnostics, task metadata, decision frequency, and the model family. Our contributions are an HDBO control formulation covering strategy selection, configuration, and execution length; the \method{}--\engine{} architecture with shared observations and HDBO-specific diagnostics; and an empirical evaluation of its performance, behavior, and inference cost.

\section{Background}
\label{sec:background}
\label{sec:related}

Given a budget of $T$ evaluations, we minimize an expensive black-box function $f:\mathcal{X}\to\mathbb{R}$ over a bounded domain $\mathcal{X}\subseteq\mathbb{R}^{D}$. Evaluating $x_t$ yields $y_t=f(x_t)+\epsilon_t$, and the resulting data $\mathcal{D}_t=\{(x_i,y_i)\}_{i=1}^{t}$ define the incumbent $y_t^\star=\min_{i\leq t}y_i$. Standard BO fits a probabilistic surrogate to $\mathcal{D}_t$ and uses an acquisition function to select $x_{t+1}$~\citep{bosurvey1,bosurvey2}.

\textbf{High-dimensional Bayesian optimization.} When $D$ is large relative to $T$, limited observations make both surrogate fitting and acquisition optimization difficult~\citep{vanilla-understanding,hdbo-survey2}.
HDBO methods impose different structural assumptions or search biases.
Variable-selection methods assume that only a subset of variables is relevant and iteratively optimize over a small set of important dimensions~\citep{dropout,saas,mcts-vs,lassobo}.
Embedding-based methods assume that the objective admits a low-dimensional subspace or embedded manifold, and perform optimization in a reduced space before projecting back to the original space~\citep{rembo,hesbo,alebo,baxus}.
Decomposition-based methods assume that the objective can be decomposed into a sum of low-dimensional functions, thereby exploiting low-order interactions or additive structures~\citep{addgpucb}.
Trust-region methods instead restrict search to a local region around the current best point~\citep{turbo}.
Carefully designed full-dimensional BO can also be competitive when such structural assumptions are absent or hard to verify~\citep{vanilla-0,vanilla-understanding}.
These are useful inductive biases for the unknown objective.
Section~\ref{sec:method} describes how \engine{} makes several such biases available to the agent; methodological details and further HDBO references appear in Appendix~\ref{app:related-work}.

\textbf{Large language models for black-box optimization.} Foundation models offer several complementary capabilities for black-box optimization: extracting priors from free-form task context, representing optimization behavior with flexible sequence models, and transferring performance prediction across search spaces~\citep{song2024foundational}. One line of work realizes these capabilities within an explicit BO loop by assigning the LLM a particular component. At the modeling stage, LLMs can design or evolve GP kernels~\citep{cake}, provide representations for probabilistic surrogates~\citep{bopro,embedregress,reelicit}, or supply low-cost predictions that are fused with real evaluations~\citep{labo}. At the decision stage, they can select acquisition functions~\citep{lmabo}, construct or filter candidates~\citep{bora,reasoningbo}, or select between LLM and BO proposals~\citep{llinbo}. LLMs can also translate feedback into observations for preference-based BO~\citep{coexbo,lilo}. These methods enrich a prescribed BO pipeline while retaining numerical modeling or acquisition machinery. A second line gives the language model a more direct role in constructing evaluations. OPRO~\citep{opro} conditions on previous solutions and objective values to generate new proposals. LLAMBO~\citep{llambo}, a principal baseline in our study, reformulates several BO components as language-model inference: it uses task descriptions for zero-shot warm-starting, samples promising configurations conditioned on the observation history, and predicts candidate performance with an LLM surrogate. LLAMBO therefore extends beyond point proposal, but its intervention interface remains fixed. Across both lines, the location and form of LLM intervention are algorithm-specified.

\textbf{Agentic BBO.} Recently, agentic systems organize optimization as a continuing reason--act--observe process and give the model authority beyond a single prescribed component. AgentHPO~\citep{agenthpo} interprets an HPO task, proposes configurations, and updates its recommendations from trial outcomes; OptiMindTune~\citep{optimindtune} distributes recommendation, evaluation, and decision making across specialized agents. These systems broaden the scope of autonomous optimization, but do not generally retain an explicit uncertainty-aware numerical backend. Hybrid systems couple agent reasoning to such machinery: Centaur~\citep{centaur} exposes CMA-ES state and proposals to an autoresearch agent, whereas Sara~\citep{sara} lets an agent inspect and reconfigure a Bayesian backend during the run. Sara is therefore the most closely related prior system. In HDBO, however, complete candidate generation is a difficult numerical task and the useful search bias is unknown. Therefore, our proposed \method{} controls which bias to test and how to configure it, while \engine{} generates several consecutive candidates before the agent reassesses the strategy. Table~\ref{tab:sara-hera} compares the primary control roles.

\begin{table}[h]
\caption{Comparison of the primary control roles in Sara with lenz~\citep{sara} and our HERA
with PRISM.}
\label{tab:sara-hera}
\centering
\begin{tabularx}{\linewidth}{@{}>{\raggedright\arraybackslash}p{0.25\linewidth}
                              >{\raggedright\arraybackslash}X
                              >{\raggedright\arraybackslash}X@{}}
\toprule
Aspect & Sara + lenz & HERA + PRISM \\
\midrule
Agent role & Experiment designer & Structure strategist \\
Decision target & Next evaluation point & Structural hypothesis \\
Candidate generation & LLM or lenz & PRISM only \\
Control level & Point-level & Strategy-level \\
Control frequency & Every evaluation & Every search block \\
Adaptation & Point and BO policy & HDBO strategy \\
Backend role & Advisory instrument & Numerical executor \\
\bottomrule
\end{tabularx}
\label{tab:HERA_Sara_Compare}
\end{table}

\section{HERA: Hypothesis- and Evidence-guided Research Agent}
\label{sec:method}

Extending Agentic BO to high dimensions changes the division of labor between the LLM and the numerical optimizer. As motivated in Section~\ref{sec:intro}, neither direct LLM candidate generation nor commitment to a single numerical backend is reliable when the objective's useful structure is unknown. The central question is therefore not which point the LLM should propose, but which structural hypothesis should guide numerical search and when that hypothesis should be revised. We assign the LLM this strategy-level role: \method{} maintains a working hypothesis and selects how it should be operationalized, while \engine{} executes the corresponding numerical strategy and generates evaluation points. This gives the LLM strategic authority without asking it to solve high-dimensional numerical optimization directly (Figure~\ref{fig:method}).

\paragraph{Agentic HDBO.}
We follow the general metalevel decision-process view of Agentic BO introduced by Sara~\citep{sara}, but define a different action space for high-dimensional search. Apart from one explicitly separated initialization action, HERA never emits a candidate point. Its admissible actions are
\begin{equation}
  \mathcal{A}_{\mathrm{HERA}}
  =\mathcal{A}_{\mathrm{state}}
  \cup\mathcal{A}_{\mathrm{init}}
  \cup\mathcal{A}_{\mathrm{strategy}}
  \cup\mathcal{A}_{\mathrm{block}}
  \cup\{\mathtt{stop}\},
  \label{eq:hera-action-space}
\end{equation}
where the first four sets correspond exactly to the four typed PRISM tools:
\begin{equation}
  \begin{aligned}
    \mathcal{A}_{\mathrm{state}}
      &=\{\mathtt{get\_state}(\nu)\}, \\
    \mathcal{A}_{\mathrm{init}}
      &=\{\mathtt{propose\_initial\_solution}(\iota)\}, \\
    \mathcal{A}_{\mathrm{strategy}}
      &=\left\{\mathtt{set\_search\_strategy}(\rho,b,\lambda,\tau,r):
        \rho\in\{\mathtt{switch},\mathtt{reconfigure},\mathtt{restart}\}\right\}, \\
    \mathcal{A}_{\mathrm{block}}
      &=\left\{\mathtt{run\_search\_block}(L,r):
        1\leq L\leq\min(50,T-t)\right\}.
  \end{aligned}
  \label{eq:hera-tool-actions}
\end{equation}
For a state query, $\nu$ specifies the state and structure detail levels, whether to repeat task metadata or the strategy catalog, and the recent window and number of top observations to return. For the one-shot initialization action, $\iota=(v,\phi,\delta,r)$ contains named variable values $v$ in the task's original representation, a normalized fill value $\phi$ for unnamed coordinates, an optional spread $\delta$, and a rationale $r$. In a strategy action, $b$ names a PRISM strategy, $\lambda$ contains its schema-validated configuration, $\tau$ optionally identifies the component to restart, and $r$ records the reason for the intervention. Arguments not used by the selected intervention $\rho$ are omitted. Finally, a block action specifies its evaluation horizon $L$ and rationale $r$; it is the only action that invokes the evaluator. The initialization action merely queues a design, which is charged when the next block evaluates it.
The \texttt{stop} action is handled by the agent loop rather than PRISM and terminates the campaign without changing numerical state. A decision cycle may therefore contain several zero-budget actions followed by at most one budget-consuming block. Numerical models update after every observation inside that block, whereas HERA reassesses the structural hypothesis only after control returns.

Accordingly, HERA's decision context augments the usual Agentic BO state with the active HDBO strategy, structural evidence, and the realized return of previous blocks. These quantities support decisions under structural uncertainty. The key difference from point-level Agentic BO is thus not merely an action set: HERA controls the structural search process, while PRISM retains authority over complete high-dimensional candidate generation.

\textbf{HDBO diagnostics and working hypotheses.} In high dimensions, the same lack of progress can result from inadequate coverage, an unreliable surrogate, or a mismatch between the active search bias and the objective. PRISM therefore augments routine optimization feedback with three HDBO-specific diagnostic groups: search geometry, summarized by normalized OTSD and observation entropy~\citep{otsd}; surrogate credibility, assessed through posterior uncertainty and leave-one-out residuals; and provisional evidence for coordinate sparsity, low-dimensional variation, and decomposable interactions. These signals are returned as bounded summaries and are aligned with the numerical strategies available to HERA. Appendix~\ref{app:diagnostic-state} provides their definitions, returned fields, and failure cases.

HERA combines these diagnostics with the active strategy, realized block improvement, and remaining budget to maintain a \emph{working hypothesis}: an explanation of which search bias may currently be useful. The hypothesis links supporting and conflicting evidence to a numerical intervention and a condition for reassessment. Surrogate credibility qualifies the structural signals, while subsequent evaluator outcomes test whether the chosen bias was useful. Evaluator returns remain observed evidence; diagnostics are derived evidence and metadata is prior context. Thus, a working hypothesis guides adaptation without claiming that HERA has recovered the objective's true structure.

\paragraph{PRISM tools.}
HERA acts through four typed PRISM tools (Table~\ref{tab:prism-tools}). These tools are exposed directly to the language model as structured function calls; HERA does not operate PRISM through a shell or a free-form CLI.

\begin{table}[htbp]
\caption{PRISM tools available to HERA, with arguments validated by PRISM.}
\label{tab:prism-tools}
\centering
\begin{tabularx}{\linewidth}{@{}l X@{}}
\toprule
Tool & Returned information or effect \\
\midrule
\texttt{get\_state} & Budget and progress, active strategy and scoreboard, geometry, surrogate credibility, structural evidence, and optional metadata/catalog. No state change. \\
\texttt{propose\_initial\_solution} & Validate and queue one complete initial design. The next search block evaluates it first. \\
\texttt{set\_search\_strategy} & Atomically switch, reconfigure, or restart the numerical strategy without an objective evaluation. \\
\texttt{run\_search\_block} & Return every observed value and best-so-far, block improvement, and remaining budget. \\
\bottomrule
\end{tabularx}
\end{table}

The formal action classes in Equations~\ref{eq:hera-action-space} and \ref{eq:hera-tool-actions} map directly to these calls. State queries are read-only and return bounded summaries rather than unrestricted backend objects. Switching selects a different inductive bias; reconfiguration changes the active strategy's semantic hyperparameters; restarting rebuilds a strategy-specific component, such as a trust region, embedding, grouping, or active set. PRISM rejects unavailable strategies, invalid configuration keys, dense candidate vectors, out-of-range block sizes, and actions that violate the one-shot initialization contract.

The numerical strategies available through this interface embody distinct assumptions about the objective. We choose representative methods for several common HDBO hypotheses: Vanilla BO for a full-dimensional default~\citep{vanilla-0}, LassoBO for axis-aligned sparsity~\citep{lassobo}, an ALEBO-inspired low-rank strategy for low-dimensional subspace structure~\citep{alebo}, Add-GP-UCB for additive interactions~\citep{addgpucb}, and TuRBO for local regularity~\citep{turbo}. The broader sparse and embedding families also include methods such as SAASBO and BAxUS~\citep{saas,baxus}. These strategies provide numerical mechanisms through which HERA tests and revises working hypotheses as evidence accumulates, rather than fixing one hypothesis for the entire run.

After the shared opening design, HERA may translate documented task semantics into one proposed design. PRISM validates it, places it at the front of the next search block, and charges it to the common evaluation budget. Once any search block has begun, this channel closes and every subsequent candidate is generated by a numerical strategy. This makes the role of semantic initialization explicit rather than conflating it with later Agentic HDBO decisions.

\paragraph{Search blocks.}
Given an accepted \texttt{run\_search\_block}$(L_k,r_k)$ action, PRISM runs the active backend sequentially for at most $L_k$ evaluations. Any strategy change is validated and applied first, before the block starts. For $t=t_k,\ldots,t_k+L_k-1$,
\begin{align}
  x_{t+1}
    &=\operatorname{Suggest}_{\mathcal{B}_k,\lambda_k}
      (\mathcal{D}_{t}), \\
  y_{t+1}
    &=f(x_{t+1})+\epsilon_{t+1},
  &
  \mathcal{D}_{t+1}
    &=\mathcal{D}_{t}\cup\{(x_{t+1},y_{t+1})\}.
  \label{eq:prism-update}
\end{align}
PRISM fits the numerical model, optimizes its acquisition rule, materializes a complete candidate, and checks its dimensionality, bounds, declared constraints, and duplication status before evaluation. The backend can update after each returned observation. All evaluated points remain in the shared observation history, so a newly selected strategy can reuse previous evidence even when incompatible strategy-specific state must be rebuilt.

The block terminates when $L_k$ evaluations have been completed, the remaining budget is exhausted, or the evaluator fails. PRISM then constructs the next context $c_{k+1}$ from the updated history and strategy state, returning control to HERA. This transition establishes the central boundary of the method: HERA decides what numerical experiment to run and how long to run it, while PRISM retains authority over ordinary high-dimensional candidate generation, model fitting, evaluation, and budget accounting.

\section{Experiments}
\label{sec:experiments}

We first evaluate whether existing methods can transfer to high-dimensional tasks, then compare HERA with simpler controllers over the same numerical backends. Real-world benchmarks and ablations assess the complete system. Our code is available in the supplemental files.

\paragraph{Experimental Setup.}
Unless stated otherwise, each run has a budget of 100 objective evaluations, including five shared scrambled-Sobol opening points, and uses 10 seeds. All objectives are minimized. We compare \method{} with Sobol search; five classical HDBO methods---Vanilla BO~\citep{vanilla-0}, TuRBO~\citep{turbo}, Add-GP-UCB~\citep{addgpucb}, ALEBO~\citep{alebo}, and LassoBO~\citep{lassobo}; and LLM-based or agentic systems---LLAMBO~\citep{llambo}, Centaur~\citep{centaur}, and Sara~\citep{sara}. The principal LLM comparison uses DeepSeek-V4.1-Flash. We report incumbent trajectories and final objectives with variability across seeds. A task-level win refers to the mean final objective, not every seed. Real-world LLM systems receive task information that classical HDBO methods do not use, so those comparisons evaluate complete systems. Appendix~\ref{app:protocol} documents task definitions, representations, and implementation details.

\subsection{Synthetic Benchmarks}
\label{sec:synthetic}

We first examine whether existing LLM-based and agentic BO methods transfer to high-dimensional problems without task metadata. The suite comprises Hartmann6 (extended to 100-D), Ackley10 (extended to 200-D), L\'evy15 (extended to 300-D), and Rosenbrock10 (extended to 300-D), covering a range of optimization challenges. Task identities are anonymized and metadata cards are empty.

\begin{figure}[t]
  \centering
  \includegraphics[width=\textwidth]{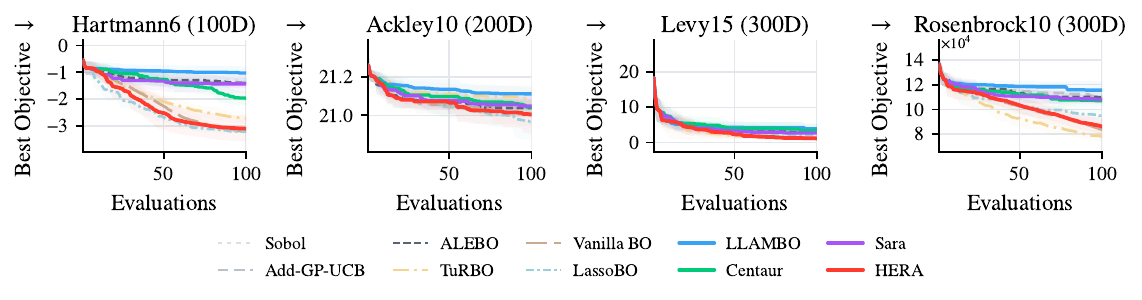}
  \caption{\textbf{Metadata-free high-dimensional optimization.} Incumbent trajectories on four synthetic tasks, comparing \method{}, classical HDBO methods, and LLM-based methods (LLAMBO~\citep{llambo}, Centaur~\citep{centaur}, Sara~\citep{sara}) over 10 seeds. The agentic baselines show a transfer gap, while \method{} remains competitive.}
  \label{fig:synthetic}
\end{figure}

As shown in Figure~\ref{fig:synthetic}, the LLM-based and agentic baselines yield worse mean final objectives than strong numerical HDBO methods across all four tasks (mean 
$ \pm $ std over 10 seeds), indicating that naively extending these methods to high-dimensional settings fails to deliver competitive performance.  In contrast, HERA performs on par with the strongest traditional high-dimensional BO baselines: it ranks first on L\'evy, second on Ackley, and third on Hartmann and Rosenbrock, matching or outperforming these numerical methods. Appendix~\ref{app:tasks} gives the details.

The variation across functions is as informative as the aggregate ranking. No single strategy wins everywhere: full-dimensional or sparse search works best on Hartmann, HERA is best on L\'evy, and local trust-region search leads on Rosenbrock. In other words, there is no one high-dimensional search bias that is always preferable. What HERA offers is therefore not a universally better strategy, but a robust one: it stays close to the best numerical methods without relying on the hidden construction of the task or any prescribed structural assumption. This is exactly what our formulation aims for—the agent does not need to recover the true structure, only to keep a useful numerical search strategy as evidence accumulates.

\subsection{Heuristic and Bandit Baselines}
\label{sec:scheduling}

We next ask how much of HERA's performance can be explained by simpler backend selection over the same numerical portfolio. The structure-aware heuristic maps the same high-dimensional diagnostics to backends through fixed rules. UCB~\citep{ucb1} instead treats each backend as an arm. Both baselines use the same portfolio, initialization, evaluation budget, and fixed five-evaluation blocks on the four synthetic tasks. HERA can additionally configure or restart a backend and choose its execution horizon. Appendix~\ref{app:scheduling} specifies the policies.

\begin{figure}[t]
  \centering
  \includegraphics[width=\textwidth]{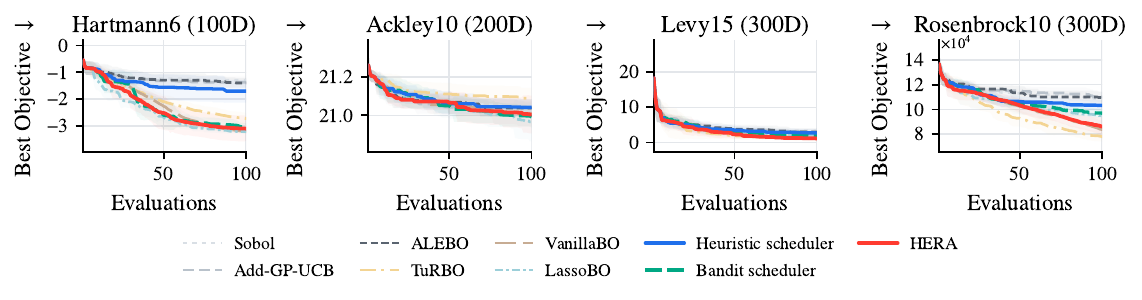}
  \caption{\textbf{Comparison with heuristic and bandit baselines.} \method{}, a structure-aware heuristic, and UCB use the same five numerical backends on four synthetic tasks. \method{} is better than both adaptive baselines on three tasks and ranks second to UCB on Ackley.}
  \label{fig:scheduling}
\end{figure}

Figure~\ref{fig:scheduling} shows that HERA performs better than both controllers on Hartmann, L\'evy, and Rosenbrock, while UCB is slightly better on Ackley. The performance of the two baselines also varies by task: UCB is competitive on Hartmann and Ackley, whereas the heuristic is consistently weaker. The scheduler traces clarify why UCB's competitiveness on two tasks does not make it a sufficient replacement for HERA. As Figure~\ref{fig:scheduler-traces} in Appendix~\ref{app:scheduling} shows, the backend allocations of UCB are nearly task-invariant: it frequently switches across the portfolio at a similar rate across the four objectives, regardless of their different structural properties. Its switching is therefore driven primarily by recent normalized improvement rather than by task-specific structural evidence. This is the essential limitation of a reward-only controller: it can react to whichever backend happened to do well last, but it cannot represent, revise, or condition on a structural hypothesis. HERA instead exhibits task-dependent allocation patterns, concentrating on different subsets of strategies across objectives.

\subsection{Real-World Benchmarks}
\label{sec:real}

\begin{figure}[t]
  \centering
  \includegraphics[width=1.0\textwidth]{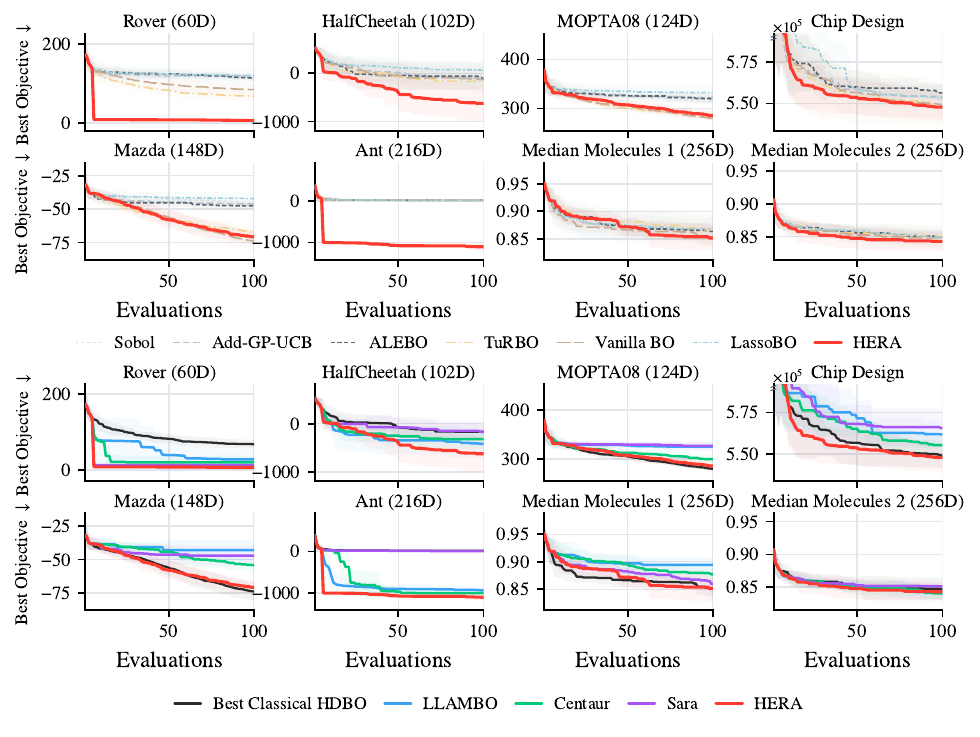}
  \caption{\textbf{Real-world tasks.} Comparisons with classical HDBO methods (top) and with LLM-based methods (bottom) over 100 evaluations and 10 seeds. Full results are reported in Appendix~\ref{app:real}.}
  \label{fig:real}
\end{figure}

We next evaluate the complete HERA system on eight real-world HDBO tasks: Rover (60-D)~\citep{rover}, HalfCheetah (102-D)~\citep{halfcheetah}, MOPTA (124-D)~\citep{saas}, Chip Design (128-D)~\citep{bboplace}, Mazda (148-D)~\citep{mazda}, Ant (216-D)~\citep{ant}, and Median Molecules 1 and 2 (256-D)~\citep{guacamol}. The suite spans trajectory design, continuous control, engineering, chip placement, and molecular optimization. HERA and the LLM-based or agentic baselines receive the same task information, whereas the classical HDBO methods operate only on evaluated inputs and outputs. Figure~\ref{fig:real} reports these two comparator groups separately; the latter comparison is therefore a complete-system evaluation rather than an isolation of language metadata.

Compared with classical HDBO methods, HERA obtains a better mean final objective than the strongest evaluated baseline on six of eight tasks. The largest improvements occur on Rover, HalfCheetah, and Ant. On MOPTA and Mazda, full-dimensional BO remains better, while HERA is competitive. On Median Molecules 1 and 2 and Chip Design, HERA has the better performance. Thus, HERA is broadly competitive across heterogeneous tasks.
Compared with the LLM-based and agentic methods, HERA achieves the best mean final objective on seven tasks. The exception is Median Molecules 2, where Centaur and LLAMBO are slightly better. The contrast is especially clear on Rover, HalfCheetah, and Ant: although several agentic methods benefit from task information, HERA attains a better final mean than each of them and also matches or exceeds the strongest classical HDBO baseline (black curve). On Chip Design, HERA also has the best mean across both comparator groups. Notably, other LLM-based and agentic methods perform worse than the strongest classical HDBO baseline on tasks like Chip Design and Median Molecules 1, whereas HERA is still competitive. Appendix~\ref{app:real} reports the complete final values and seed variability.

Taken together, the results support HERA as a strong complete system for real-world Agentic HDBO. The largest gains appear on Rover, HalfCheetah, and Ant, where semantic context can inform initialization and numerical feedback remains important during search. Smaller gains on the engineering and molecular tasks show that an effective fixed optimizer can remain difficult to improve upon. Because HERA combines task semantics and online strategy adaptation, these experiments establish the value of the complete framework but do not attribute the gains to backend orchestration alone; the following ablations examine these factors separately.

\subsection{Ablation Studies}
\label{sec:evidence}

Having established end-to-end performance, we next ask which information and control choices shape HERA's behavior, performance, and cost. We study structural diagnostics, task metadata, decision frequency, and agent model. Detailed trajectories and statistics are deferred to the appendix.

\paragraph{Structural evidence.}
We first ask whether the HDBO-specific diagnostics provide a meaningful control signal. We remove the sparsity, low-dimensional-subspace, and interaction channels while retaining the other components. Removal alters the strategies adopted and makes backend use substantially more concentrated, most visibly toward Vanilla BO on MOPTA and Mazda (Appendix Figure~\ref{fig:ablation-timeline}). Final performance, however, remains similar on most tasks, so the experiment supports a behavioral effect rather than an aggregate optimization gain. An offline audit reinforces this interpretation: variable-relevance and low-dimensional-subspace readings become more informative as observations accumulate, whereas the interaction-grouping diagnostic remains unreliable. These diagnostics should therefore be viewed as provisional evidence for deciding which search bias to test, not as recovered ground-truth structure; Appendices~\ref{app:evidence} and~\ref{app:diagnostics} provide the full results.

\paragraph{Task metadata.}
We next test whether semantic task information contributes beyond numerical observations. The strict-removal condition withholds task identity, descriptions, and the corresponding prompt instructions while preserving the optimization budget and interface. Removal degrades performance on most tasks, with the clearest effects on Rover and Ant, but slightly improves performance on a few other tasks (Appendix~\ref{app:metadata}). It also almost eliminates the use of the initial-solution tool. Metadata therefore acts as a task-dependent prior rather than a universally beneficial input, and its effect can enter through both initialization and later reasoning. This intervention consequently does not isolate metadata-driven strategy selection from semantic initialization.

\paragraph{Decision frequency.}
We then ask whether HERA should reconsider its strategy after every evaluation. On HalfCheetah, MOPTA, and Mazda, we compare adaptive blocks with blocks of size one, both with and without context compaction. With context compaction, returning control after every evaluation increases inference by roughly an order of magnitude without a consistent improvement across paired runs. Without context compaction, making decisions after every evaluation helps on some tasks, but the improvement is not consistent across tasks (Figure~\ref{fig:cost}). Adaptive reassessment is therefore the more practical default: it allows the numerical backend to accumulate evidence before deliberation while avoiding a large inference overhead.

\paragraph{Model family.}
To measure sensitivity to the underlying LLM, we replace DeepSeek-V4.1-Flash with Kimi-K3 and GPT-6-astra on the same three tasks. Kimi-K3 has the best final mean on HalfCheetah, whereas GPT-6-astra has the best means on MOPTA and Mazda (Appendix~\ref{app:models}). No model dominates across tasks. Agent-model choice is therefore another task-dependent design factor rather than a uniformly improving substitution.

Additional analyses of tool-use trajectories and token consumption are provided in Appendices~\ref{app:tool-use} and~\ref{app:tokens}. They show selective, task-dependent use of the backend portfolio, but remain descriptive and do not establish that more frequent interventions cause better performance.
Collectively, the ablations show that HERA's information and control interface materially changes how optimization is conducted. Structural evidence primarily changes strategy use, metadata can change both initialization and subsequent decisions, and per-evaluation deliberation sharply increases inference cost. Model rankings and metadata effects remain task dependent.

\section{Conclusion}

We presented \method{} and \engine{}, an agent--engine framework for high-dimensional Bayesian optimization in which an LLM manages revisable structural hypotheses and search strategies while numerical backends generate high-dimensional candidates. On metadata-free synthetic functions, \method{} remains competitive with strong HDBO methods and improves over the evaluated LLM-based and agentic baselines; across eight real-world tasks, it performs strongly against other methods. The ablations further show that structural evidence changes strategy use, metadata provides task-dependent priors, and adaptive decision intervals control inference cost. Together, these results support a view of Agentic HDBO in which the LLM manages the evolving search strategy rather than directly solving high-dimensional candidate generation.
The current framework is bounded by the diagnostics and numerical strategies exposed by \engine{}: \method{} can test and revise hypotheses within this interface, but its working hypotheses do not guarantee recovery of the objective's true structure. Future work could expand the available tool and strategy space, calibrate the reliability of structural diagnostics, and transfer strategy experience across related optimization tasks.

\newpage
\subsection*{AI use statement}
In this work, we used generative AI tools for translation and language-related assistance, including improving grammar, clarity, concision, and the presentation of author-written text. We have not used generative AI tools for formulating research questions or hypotheses, developing the conceptual or theoretical framework, designing the methodology or experiments, implementing the proposed method, generating or processing data, or analyzing and interpreting the experimental results. We have reviewed all AI-assisted work. All AI-assisted revisions were checked against the authors' intended meaning and reviewed for technical accuracy. We take responsibility for the final content of this work, including text, claims or artifacts produced with the aid of generative AI.

\subsubsection*{Reproducibility statement}
The appendices specify task representations, opening designs, evaluation budgets, backend adaptations, scheduler rules, intervention scopes, and sample counts. All objective evaluations, including prior proposals, count toward the same budget. No additional results are inferred from unexecuted experiments. The code is provided in supplementary material.

\bibliography{iclr2027_conference}
\bibliographystyle{iclr2027_conference}

\clearpage
\appendix
\section{Further Related Work}
\label{app:related-work}

\paragraph{Classical High-dimensional Bayesian Optimization.}
Random embeddings such as REMBO and HeSBO search through lower-dimensional representations~\citep{rembo,hesbo}. ALEBO modifies embedding geometry, whereas BAxUS expands nested subspaces as the search progresses~\citep{alebo,baxus}. Other approaches fit sparse-coordinate models or select active variables~\citep{saas,mcts-vs,lassobo}, model additive components with overlapping, tree-structured, or randomized decompositions~\citep{addgpucb,addoverlapping1,addtreestructure,rducb}, or search with adaptive local trust regions~\citep{turbo}. Full-dimensional BO with dimension-aware priors and careful candidate generation remains a strong alternative~\citep{vanilla-0,vanilla-1,vanilla-understanding}. These lines of work supply different biases.

\paragraph{Language models at different BO interfaces.}
Methods with an explicit numerical BO loop use language models at several interfaces. CAKE uses LLM operations to generate and refine GP kernels~\citep{cake}. Embed-then-Regress uses fixed language-model embeddings, GOLLuM jointly adapts an LLM representation with a GP through marginal likelihood, and ReElicit repeatedly constructs a semantic representation for surrogate modeling~\citep{embedregress,gollum,reelicit}. LABO treats LLM predictions as a low-cost information source and fuses them with experimental observations in a multi-fidelity model~\citep{labo}. At the decision layer, LMABO selects an acquisition function~\citep{lmabo}; BOPRO, BORA, and Reasoning BO condition, propose, or rank candidates~\citep{bopro,bora,reasoningbo}; LLINBO arbitrates between LLM and BO proposals~\citep{llinbo}; and LILO maps natural-language feedback to preference observations~\citep{lilo}. A separate family places proposal generation more directly in the language model: OPRO generates solutions from the optimization trajectory, LLAMBO combines language-model prediction with promising-point generation, and HOLLM augments iterative LLM optimization with search-space partitioning~\citep{opro,llambo,hollm}. These approaches differ in how much numerical BO machinery they retain, but each prescribes the interface through which the LLM affects search.

\paragraph{Agentic control and adaptive numerical policies.}
Agentic optimization spans several levels of control. AgentHPO iteratively proposes hyperparameters from task descriptions and trial outcomes and OptiMindTune divides HPO among cooperating agents~\citep{agenthpo,optimindtune}. Centaur exposes CMA-ES progress to an autoresearch agent and combines numerical and LLM proposals~\citep{centaur}. Sara gives an agent access to a Bayesian backend, including surrogate inspection, evaluation selection, and run-time policy changes~\citep{sara}. Its scope of control makes it the closest agentic BO comparison, while the high-dimensional setting motivates our block-level control interface.

\section{Experimental Settings and Task Definitions}
\label{app:protocol}

\subsection{Tasks and representations}
\label{app:tasks}

Table~\ref{tab:tasks} lists the optimization suite. Unless an ablation specifies otherwise, each task--method pair uses seeds 0--9, a total budget of 100, and a shared five-point scrambled-Sobol opening design. Numerical optimization uses normalized coordinates, converted to the native representation before evaluation. Reward-maximization tasks are oriented toward minimization. The molecular tasks optimize a continuous latent representation of the GuacaMol median-molecule objectives~\citep{guacamol}; the reported dimension is that representation's dimension, not the number of atoms or a native discrete search-space dimension.

\begin{table}[htbp]
\centering
\caption{Optimization tasks.}
\label{tab:tasks}
\begin{tabularx}{\linewidth}{@{}l r X@{}}
\toprule
Task & $D$ & Representation or construction \\
\midrule
Hartmann-100 & 100 & Six latent inputs formed by a seeded block-average embedding. \\
Ackley-200 & 200 & Dense orthogonal mixing over all coordinates; shifted coordinates. \\
L\'evy-300 & 300 & Fifteen block-average latent inputs; shifted coordinates. \\
Rosenbrock-300 & 300 & Full 300-coordinate chained Rosenbrock; shifted coordinates. \\
\midrule
Rover & 60 & Trajectory-design parameters. \\
HalfCheetah & 102 & Continuous-control policy parameters. \\
MOPTA & 124 & Engineering-design benchmark. \\
Chip Design & 128 & Coordinates of 64 macros. \\
Mazda & 148 & Engineering-design benchmark. \\
Ant & 216 & Continuous-control policy parameters. \\
Median Molecules 1 & 256 & Continuous molecular latent representation. \\
Median Molecules 2 & 256 & Continuous molecular latent representation. \\
\bottomrule
\end{tabularx}
\end{table}

The synthetic embeddings are not sparse-coordinate embeddings: all ambient coordinates contribute through block averages. Shifted tasks apply a fixed coordinatewise shift with wrap-around, $(x+\delta)\bmod 1$, before evaluating the underlying construction.

Anonymized synthetic runs expose opaque identifiers and empty metadata cards. This reduces access to named benchmark formulas, but does not guarantee that an agent cannot infer properties from observations. The optional initial-solution tool remains available.

\subsection{Baselines and Implementation Details}

Fixed numerical comparators comprise the five portfolio members and Sobol search. The strongest classical comparator is selected for each task by mean final objective. This is an evaluation reference, not a deployable oracle or an additional numerical method.

LLAMBO, Centaur, and Sara are evaluated under common task wrappers, budgets, and opening designs. The principal real-world comparison uses the same DeepSeek-V4.1-Flash endpoint and archived task-card condition.
The main \method{} configuration uses DeepSeek-V4.1-Flash at temperature zero. Model ablations use Kimi-K3 at temperature one, required by that endpoint, and GPT-6-astra at temperature zero through a separate relay.

\section{Tool details}
\label{app:tools-contract}

\begin{table}[htbp]
\centering
\caption{PRISM tools exposed to HERA and their objective-budget effects.}
\label{tab:tools}
\begin{tabularx}{\linewidth}{@{}l X l@{}}
\toprule
Tool & Contract & Budget \\
\midrule
\texttt{get\_state} & Inspect a bounded state view. Summary mode reuses compact evidence; full mode refreshes surrogate credibility and structure diagnostics. & None \\
\texttt{propose\_initial\_solution} & Propose one prior-informed design after the shared opening design. & One when evaluated \\
\texttt{set\_search\_strategy} & Atomically switch, reconfigure, or restart a strategy. Arguments are checked against the selected strategy's semantic schema. & None \\
\texttt{run\_search\_block} & Execute the requested number of sequential proposals, validate every candidate, call the evaluator, and update the shared observation history. & Evaluations returned \\
\bottomrule
\end{tabularx}
\end{table}

HERA calls these operations through the in-process \texttt{PrismToolbox} function interface. PRISM also provides command-line commands for human inspection and reproducibility, but the agent does not issue shell commands or parse terminal output.
The observed dataset is authoritative; diagnostics, scoreboards, and hypotheses are derived views. Switching starts a new backend-specific episode while preserving the complete observation history.

\section{Synthetic Results and Scheduler Definitions}
\label{app:scheduling}

\subsection{Heuristic and bandit policies}

The heuristic begins with full-dimensional BO (Vanilla BO) and reassesses after each five-evaluation block. It activates sparse search when active fraction is below $0.1$, the low-dimensional subspace backend when dimension ratio is at most $0.02$, grouped-GP search when the partition score is at least $0.9$, and local search when normalized observation-tour length exceeds $0.05$. Multiple triggers are resolved by a seeded random choice; no trigger retains the active backend. The geometry rule reflects the search distribution, not a statistical test of local smoothness.

UCB treats each block as one play. After selecting each untried backend, it maximizes
\begin{equation}
 \bar r_i+\sqrt{\frac{2\log N}{n_i}},\qquad
 r_k=\frac{y^\star_{t_k}-y^\star_{t_{k+1}}}{s(\mathcal{D}_{t_k})},
 \label{eq:ucb}
\end{equation}
where $n_i$ is the backend's play count and $N$ the total number of plays. The response scale $s$ uses $1.4826$ times median absolute deviation, with interquartile-range and standard-deviation fallbacks. This makes the bonus comparable across objective scales.

\begin{figure}[htbp]
\centering
\includegraphics[width=\textwidth]{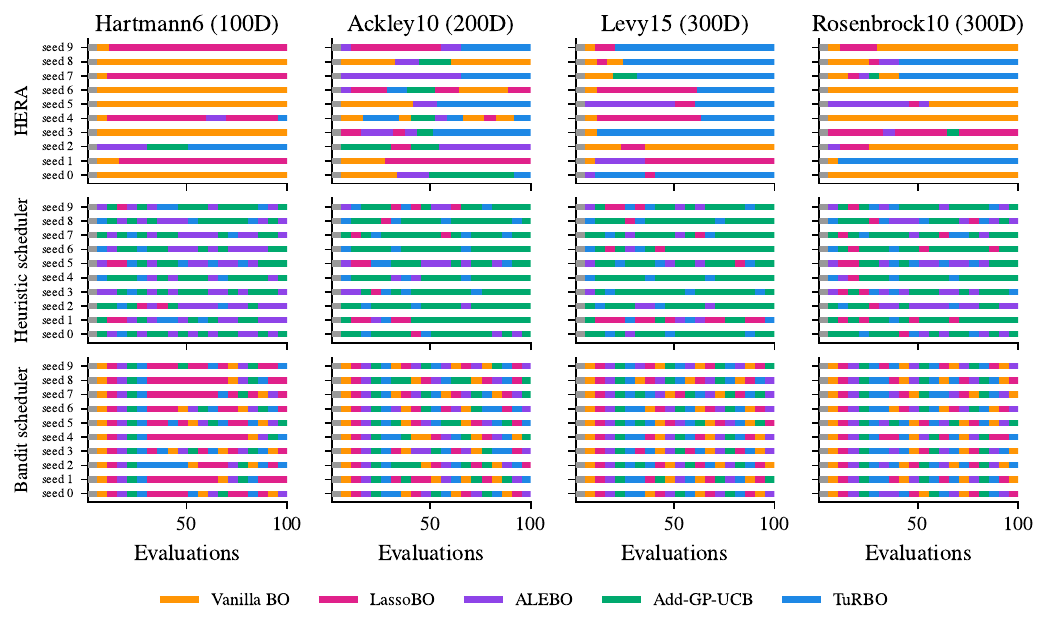}
\caption{\textbf{Scheduler trajectories and allocations.} Per-seed timelines compare \method{}, heuristic routing, and UCB. The heuristic's saturated partition-score trigger frequently allocates evaluations to grouped-GP search.}
\label{fig:scheduler-traces}
\end{figure}

Figure~\ref{fig:scheduler-traces} shows the resulting per-seed timelines and
backend allocations. The heuristic's partition-score trigger saturates and
frequently routes evaluations to Add-GP-UCB, which explains a weakness
of this particular rule. UCB is a stronger baseline and its trajectories are
close to \method{} on several tasks, but its allocations remain largely
task-invariant and are driven by recent normalized improvement rather than by
structural evidence. \method{} instead exhibits task-dependent allocation
patterns, committing longer blocks to a smaller subset of backends. These
measurements characterize the tested policies rather than learned portfolio
selection in general.

\section{Detailed Real-World Results}
\label{app:real}

Tables~\ref{tab:real-results}--\ref{tab:real-llm} report the final objectives. MM1 and MM2 abbreviate Median Molecules 1 and 2, and HC abbreviates HalfCheetah. 
Full-dimensional BO outperforms \method{} on MOPTA and Mazda. On MM2, both Centaur and LLAMBO have better means than \method{}, even though \method{} beats every fixed numerical comparator. On Chip Design, \method{} has the best mean, but its margin over full-dimensional BO is small relative to seed variation. The small MM1 gap likewise does not support a claim of substantial improvement. Large control-task gains coexist with substantial variability, particularly on HalfCheetah.

\begin{table}[htbp]
\centering
\caption{Real-world final objectives: \method{} and strong numerical comparators, mean $\pm$ sample SD over ten seeds. Lower is better.}
\label{tab:real-results}
\begin{tabular}{@{}l r r r@{}}
\toprule
Task & \method{} & Full-dimensional BO & TuRBO \\
\midrule
Rover & $6.14\pm2.04$ & $84.03\pm14.23$ & $67.64\pm12.16$ \\
HalfCheetah & $-621.40\pm370.83$ & $-135.37\pm204.38$ & $-166.72\pm166.04$ \\
MOPTA & $285.57\pm11.58$ & $280.01\pm3.97$ & $286.82\pm8.52$ \\
Mazda & $-70.73\pm5.42$ & $-73.85\pm6.88$ & $-67.08\pm5.30$ \\
Ant & $-1105.20\pm85.48$ & $16.85\pm3.20$ & $19.93\pm3.48$ \\
MM1 & $0.8514\pm0.0224$ & $0.8521\pm0.0152$ & $0.8663\pm0.0271$ \\
MM2 & $0.8430\pm0.0137$ & $0.8483\pm0.0105$ & $0.8507\pm0.0174$ \\
Chip ($10^5$) & $5.477\pm0.085$ & $5.480\pm0.061$ & $5.483\pm0.114$ \\
\bottomrule
\end{tabular}
\end{table}

\begin{table}[htbp]
\centering
\caption{Real-world final objectives for Sobol and LLM/agentic comparators, mean $\pm$ sample SD over ten seeds.}
\label{tab:real-llm}
\begin{tabular}{@{}l r r r r@{}}
\toprule
Task & Sobol & LLAMBO & Centaur & Sara \\
\midrule
Rover & $111.50\pm6.19$ & $27.46\pm24.94$ & $19.64\pm16.46$ & $12.65\pm0.32$ \\
HC & $-145.60\pm265.58$ & $-415.66\pm317.55$ & $-318.77\pm199.42$ & $-166.03\pm224.58$ \\
MOPTA & $316.38\pm5.95$ & $325.25\pm8.63$ & $299.84\pm8.50$ & $326.50\pm8.76$ \\
Mazda & $-46.30\pm2.63$ & $-42.86\pm7.36$ & $-54.17\pm5.05$ & $-47.07\pm7.27$ \\
Ant & $19.63\pm1.91$ & $-939.50\pm91.13$ & $-1007.07\pm31.16$ & $7.39\pm23.50$ \\
MM1 & $0.8644\pm0.0130$ & $0.8945\pm0.0140$ & $0.8770\pm0.0164$ & $0.8596\pm0.0238$ \\
MM2 & $0.8469\pm0.0163$ & $0.8421\pm0.0112$ & $0.8397\pm0.0070$ & $0.8513\pm0.0071$ \\
Chip ($10^5$) & $5.550\pm0.055$ & $5.617\pm0.083$ & $5.552\pm0.072$ & $5.654\pm0.113$ \\
\bottomrule
\end{tabular}
\end{table}

\section{Structural-Evidence Intervention}
\label{app:evidence}

The intervention removes the variable-sparsity, low-dimensional-subspace, and additivity entries from the explicit evidence package. Task cards, tools, opening designs, and the main prompt remain available. Backend-reported state is retained, so the intervention does not remove every possible source of structural information.

\begin{figure}[htbp]
\centering
\includegraphics[width=\textwidth]{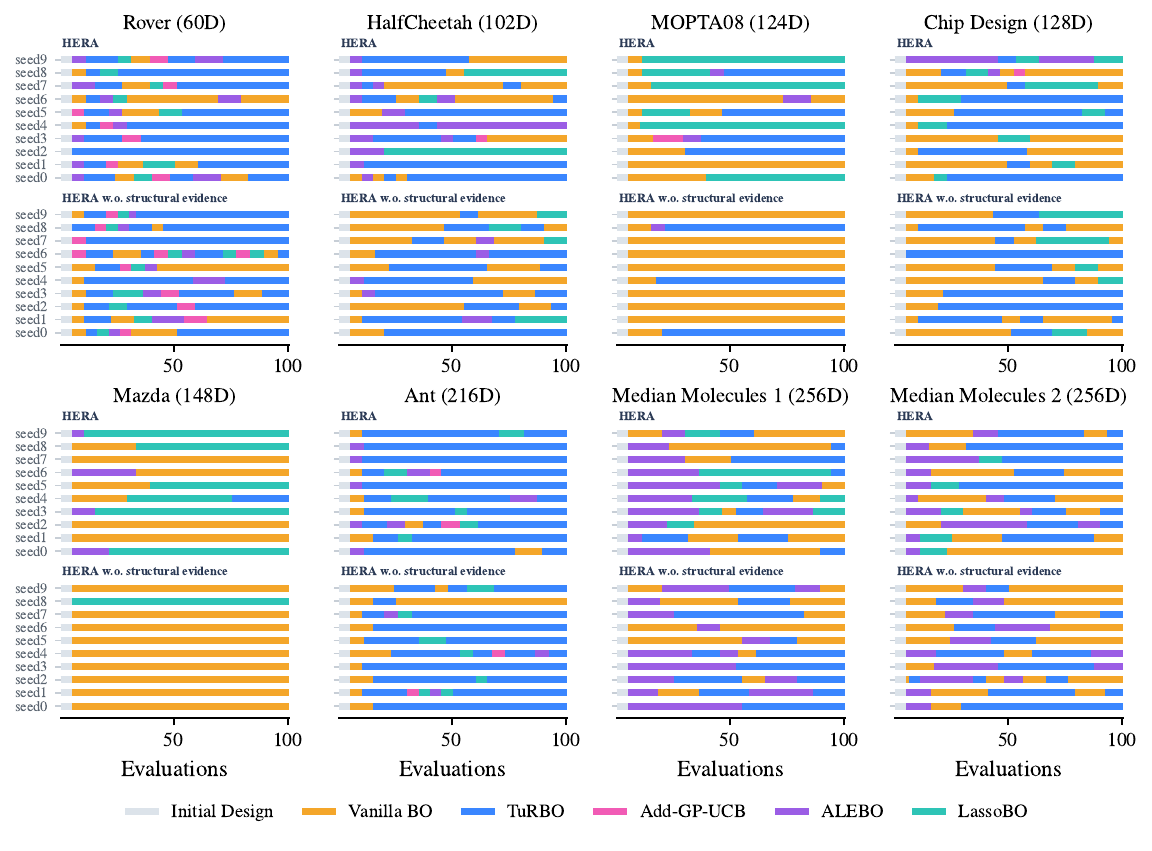}
\caption{\textbf{Structural evidence changes strategy use.} Each row is one seed and each colored segment is a backend episode over the 100 evaluations. The upper block uses full evidence and the lower block withholds the three explicit structural evidence channels.}
\label{fig:ablation-timeline}
\end{figure}

Figure~\ref{fig:ablation-timeline} shows how the intervention changes backend use across tasks and seeds. Each row traces one run over the 100-evaluation budget; the upper and lower blocks respectively show the full-evidence and removal conditions. Withholding the three structural channels produces more concentrated strategy use on several tasks, most visibly through longer Vanilla BO episodes on MOPTA and Mazda. The supported conclusion concerns control behavior: backend diversity and switching are not direct measurements of hypothesis correctness. Evidence removal may also change initialization or earlier actions, so a later trajectory difference cannot be assigned to an isolated switch.

\begin{figure}[htbp]
\centering
\includegraphics[width=\textwidth]{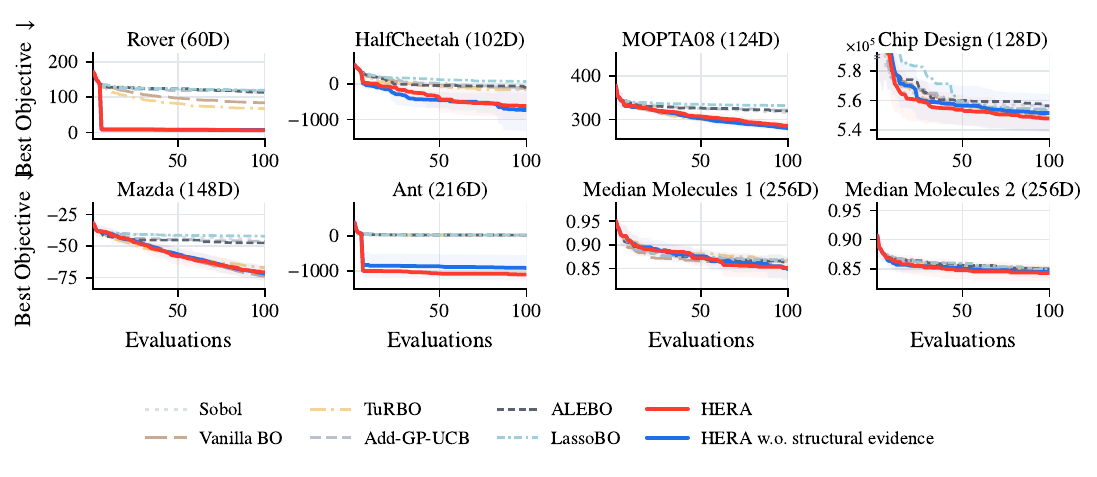}
\caption{\textbf{Final performance with and without explicit structural diagnostics.} Mean best-so-far trajectories on the eight real-world tasks over ten seeds. The curves are difficult to distinguish on most tasks, consistent with the absence of a clear aggregate performance effect.}
\label{fig:ablation}
\end{figure}

\section{Reliability of Structural Diagnostics}
\label{app:diagnostics}

\subsection{HDBO diagnostic state}
\label{app:diagnostic-state}

PRISM derives its HDBO diagnostic state from the normalized observation matrix and objective values; no diagnostic spends an objective evaluation. The interface is tailored to three questions that are coupled in HDBO: whether the observed design covers the ambient space, whether a surrogate fitted with $n\ll D$ is credible enough to interpret, and which structural bias is worth testing next. Table~\ref{tab:diagnostic-state} summarizes the quantities exposed to HERA.

\begin{table}[htbp]
\caption{\textbf{HDBO-specific diagnostic state exposed by \texttt{get\_state}.} All quantities are derived signals for controlling search, not observations of the unknown objective structure.}
\label{tab:diagnostic-state}
\centering
\begin{tabularx}{\linewidth}{@{}>{\raggedright\arraybackslash}p{0.18\linewidth} >{\raggedright\arraybackslash}X@{}}
\toprule
Channel & Principal quantities \\
\midrule
Exploration geometry & Normalized OTSD, recent OTSD gain, observation entropy, and their trajectories. \\
\midrule
Surrogate credibility & Posterior standard-deviation summaries over Sobol probes; held-out RMSE, standardized residuals, $1\sigma/2\sigma$ coverage, and $n/D$. \\
\midrule
Variable sparsity & LassoBO active-set size and fraction, selected coordinates, and a thresholded provisional verdict. \\
\midrule
Low-dimensional subspaces & Surrogate-gradient energy dimension and its ratio to $\min(D,M)$ over $M=64$ probe points. \\
\midrule
Additive structure & Interaction-graph fragmentation, number and size of candidate groups, linked pairs, and strongest normalized mixed differences. \\
\bottomrule
\end{tabularx}
\end{table}

For geometry, let $L_t$ denote the cheapest-insertion tour length through the first $t$ normalized observations. PRISM reports
\begin{equation}
  \widetilde L_t=
  \frac{L_t}{2\sqrt{5D}\,(1.5t)^{1-1/D}},
  \label{eq:normalized-otsd}
\end{equation}
as in ~\citep{otsd}. Observation entropy is the cumulative $k$-nearest-neighbor estimator used by the same exploration diagnostic. The two signals are complementary but purely geometric: a coordinate that does not affect the objective can still contribute to both distances and entropy.

Surrogate credibility is evaluated before structural readings are interpreted. PRISM fits the diagnostic GP with normalized inputs and standardized outcomes, summarizes its posterior standard deviation on 256 scrambled-Sobol probes, and performs leave-one-out-style validation. All observations are held out once when $n\leq12$; otherwise at most 12 deterministically spaced folds are used. For a held-out observation, the standardized residual is
\begin{equation}
  r_i=\frac{y_i-\widehat\mu_{-i}(x_i)}
  {\max\{\widehat\sigma_{-i}(x_i),10^{-12}\}}.
  \label{eq:loo-residual}
\end{equation}
The returned mean and standard deviation of $r_i$, empirical $1\sigma$ and $2\sigma$ coverage, and RMSE normalized by the observed response standard deviation expose bias, miscalibration, and poor generalization separately. The state also attaches $n/D$ to every structural block so that a weak fit in the undersampled regime is not misread as evidence for or against a particular structure.

\begin{figure}[htbp]
\centering
\includegraphics[width=\textwidth]{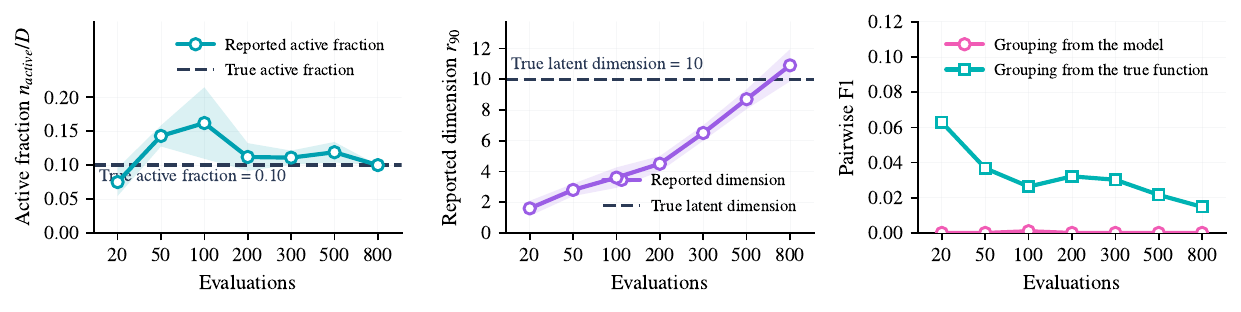}
\caption{\textbf{Offline audit of structural diagnostics.} The panels report the sparsity reading, surrogate-gradient energy dimension, and interaction-group recovery metrics used to assess the diagnostics across observation budgets.}
\label{fig:diagnostic-recovery}
\end{figure}

Figure~\ref{fig:diagnostic-recovery} summarizes the three structural readings and the corresponding offline recovery metrics used in the audit.

\subsection{Sparsity and low-dimensional variation}

Support recovery uses Jaccard overlap,
\begin{equation}
 J(\widehat S,S)=\frac{|\widehat S\cap S|}{|\widehat S\cup S|}.
\end{equation}
Given an estimated active set $\widehat S$ and, in the offline audit, a known support $S$, this score measures coordinate-support recovery. The active fraction is the size of the estimated support divided by the ambient dimension; it is a compact sparsity reading for strategy selection, not a guarantee that the selected coordinates are correct.

The low-dimensional subspace diagnostic follows gradient-based active-subspace analysis~\citep{subspace} and uses gradients of the surrogate posterior mean at probe locations to measure whether response variation is concentrated in a small number of directions. For $C=M^{-1}\sum_{j=1}^{M}g_jg_j^\top$ with eigenvalues $\lambda_1\geq\cdots\geq\lambda_D$, its energy dimension is
\begin{equation}
 d_{0.9}=\min\left\{r:\frac{\sum_{i=1}^{r}\lambda_i}
 {\sum_{i=1}^{D}\lambda_i}\geq0.9\right\}.
\end{equation}
The principal configuration uses $M=64$ probes. This statistic is an effective energy dimension of the surrogate gradient field and therefore a diagnostic for low-dimensional subspace structure, rather than an exact algebraic dimension or a recovery of the underlying subspace basis. Because it is computed from a fitted surrogate and a finite probe set, it should be interpreted together with the surrogate-credibility diagnostics.

\subsection{Interaction grouping and fragmentation}

The Differential Grouping-inspired diagnostic~\citep{differentialgrouping} computes mixed differences of the surrogate mean under coordinate perturbations, normalizes their magnitude, aggregates over anchors, and builds connected components from detected links. For component sizes $|G_1|,\ldots,|G_K|$, its reported score is
\begin{equation}
 S_{\mathrm{frag}}=\frac{D^2-\sum_{j=1}^{K}|G_j|^2}{D(D-1)}.
 \label{eq:fragmentation}
\end{equation}
This is the fraction of coordinate pairs placed in different components. It equals one for all singleton components and zero for one component. A high value can therefore reflect missed interactions, not additive structure.

\section{Metadata}
\label{app:metadata}
The main metadata ablation uses a single \emph{strict-removal} condition. It withholds task identity, semantic descriptions, reference-design hints, and the corresponding prompt instructions while preserving the optimization budget and interface. This intervention is broader than simply deleting one semantic card and is the condition reported in the main text.

\begin{figure}[htbp]
\centering
\includegraphics[width=\textwidth]{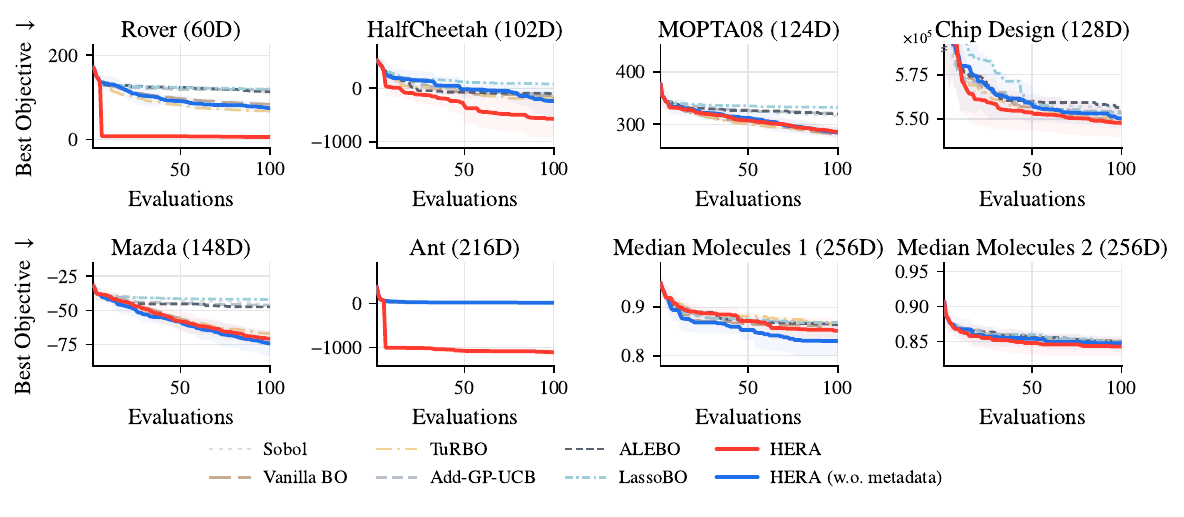}
\caption{\textbf{Metadata affects initialization and search.} Full-context and strict-removal curve. Effects are heterogeneous and do not isolate metadata-driven routing.}
\label{fig:metadata-full}
\end{figure}

As shown in Figure~\ref{fig:metadata-full}, strict removal generally worsens performance on several tasks, with the largest effects concentrated in particular tasks. The gains and reversals support task dependence rather than a uniform semantic advantage.

\section{Agent Model Configurations}
\label{app:models}

We replace DeepSeek-V4.1-Flash with Kimi-K3 and GPT-6-astra on three tasks, with ten completed seeds per configuration and the same interface and evaluation budget. Kimi's required temperature is one rather than zero, and GPT uses a separate relay. This evaluates deployed configurations rather than an isolated scalar notion of model capability.

\begin{table}[htbp]
\centering
\caption{Agent models: mean final objective $\pm$ sample SD over ten seeds. The best mean among models differs by task.}
\label{tab:models}
\begin{tabular}{@{}l r r r@{}}
\toprule
Model & HalfCheetah & MOPTA & Mazda \\
\midrule
DeepSeek-V4.1-Flash & $-621.4\pm370.8$ & $285.57\pm11.58$ & $-70.73\pm5.42$ \\
Kimi-K3 & $-706.0\pm485.1$ & $285.20\pm9.27$ & $-74.51\pm5.54$ \\
GPT-6-astra & $-323.0\pm474.2$ & $281.72\pm8.78$ & $-75.36\pm7.23$ \\
\bottomrule
\end{tabular}
\end{table}

Kimi has the best mean on HalfCheetah, whereas GPT has the best means on MOPTA and Mazda. Median ordering differs: DeepSeek's HalfCheetah median is better than Kimi's. Substantial seed variation prevents a uniform capability ranking. Model changes also shift backend usage, but behavior alone does not explain the final-objective changes.

\begin{figure}[htbp]
\centering
\includegraphics[width=\textwidth]{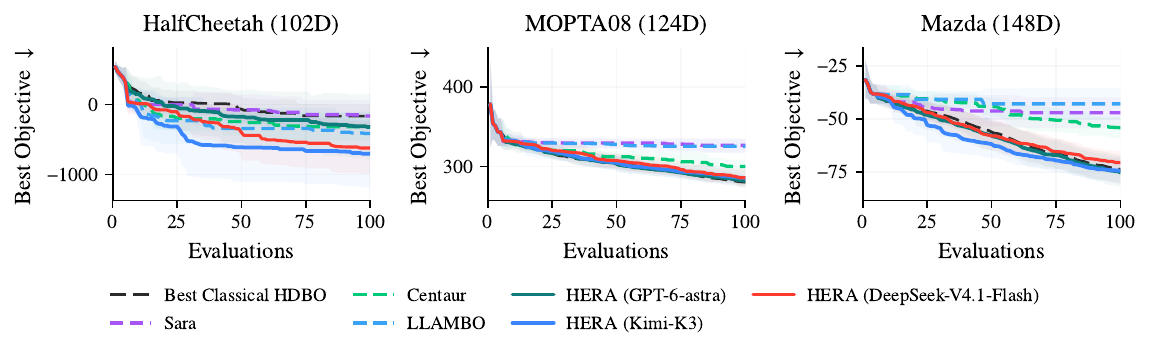}
\caption{\textbf{Model sensitivity.} Ten-seed trajectories for three agent configurations.}
\label{fig:models}
\end{figure}

\section{Tool-Use Trajectories}
\label{app:tool-use}

\begin{figure}[htbp]
\centering
\includegraphics[width=\textwidth]{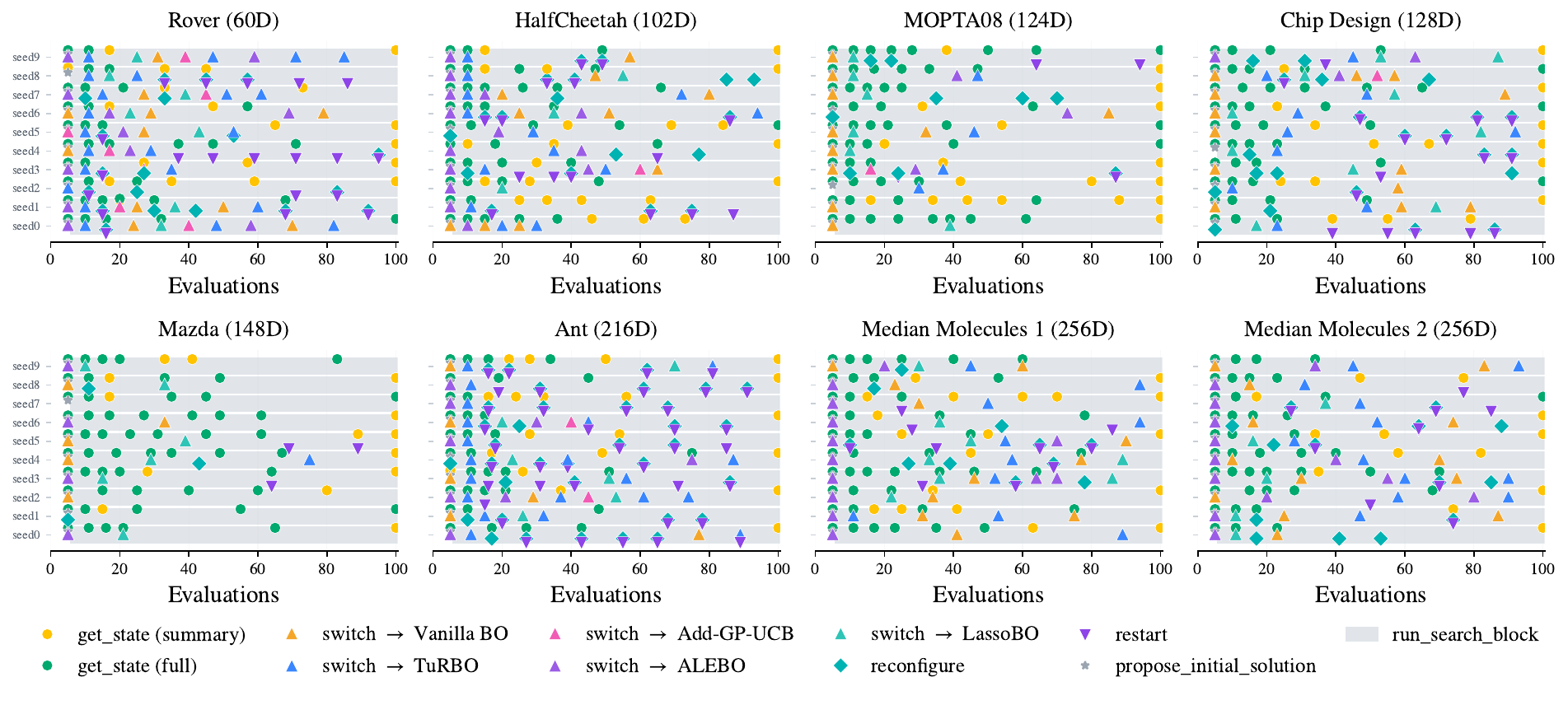}
\caption{\textbf{Tool use over the evaluation budget.} One row per campaign, grouped by task. Markers identify queries and strategy interventions; bands show evaluations consumed by blocks. A common opening protocol is followed by task- and seed-dependent behavior.}
\label{fig:tool-trajectories}
\end{figure}

The tool-use trajectories show a common initialization pattern followed by task- and seed-dependent control: some tasks trigger frequent strategy interventions, whereas others sustain a stable backend over extended search episodes (Figure~\ref{fig:tool-trajectories}). This pattern is consistent with adaptive outer-loop control, but the trajectories alone do not establish hypothesis correctness or a causal benefit from more frequent tool use.

Tool calls, assistant responses, and provider requests are distinct accounting levels, so their timelines should be read as descriptions of intervention timing rather than evidence that a hypothesis was correct or that more frequent tool use caused better outcomes.

\section{Block Size and Context Compaction}
\label{app:blocks}

The experiment compares adaptive decision blocks with and without context compaction, together with block-size-one execution under compaction. An additional uncompacted block-size-one arm provides an exploratory matched-subset comparison. Compaction keeps recent tool results detailed, summarizes older results, and preserves assistant reasoning; block size one returns after each evaluation.

\begin{figure}[htbp]
\centering
\includegraphics[width=0.58\textwidth]{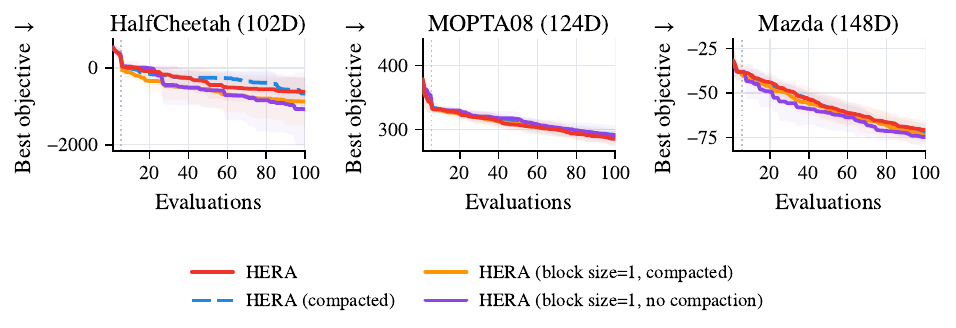}
\includegraphics[width=0.4\textwidth]{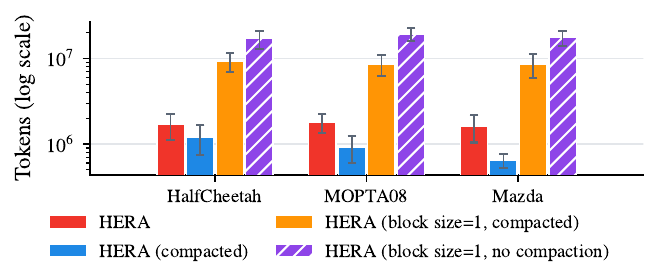}
\caption{\textbf{Performance and inference cost under different decision frequencies.} Adaptive blocks and block-size-one variants are compared with and without context compaction.}
\label{fig:cost}
\end{figure}

Figure~\ref{fig:cost} shows that context compaction substantially reduces inference cost while preserving broadly comparable optimization behavior. More frequent decisions can improve individual task trajectories, but their additional cost and the limited exploratory comparison do not support a universally best decision interval. The result is therefore a task-dependent trade-off between control frequency and inference efficiency.

\section{Token Accounting}
\label{app:tokens}

\begin{figure}[htbp]
\centering
\includegraphics[width=\textwidth]{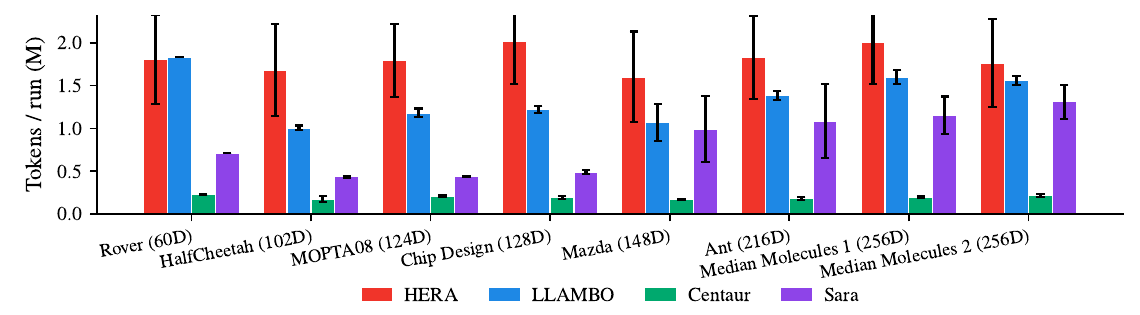}
\includegraphics[width=\textwidth]{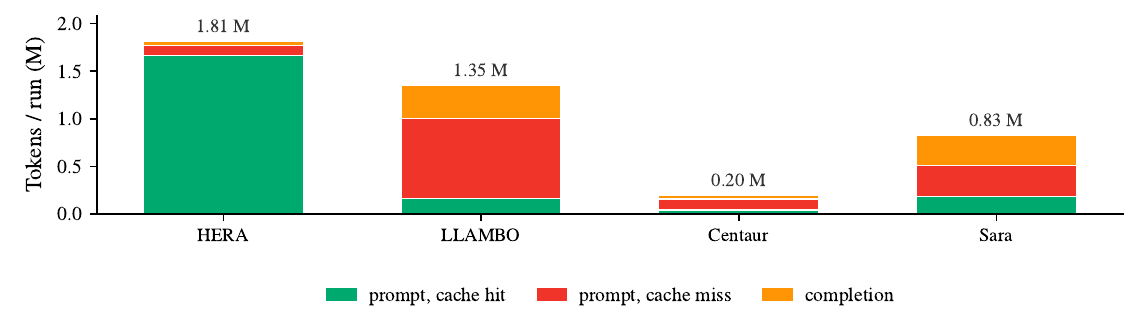}
\caption{\textbf{Tokens and cache decomposition.} Per-task totals and cached-input, uncached-input, and output components. Total-token efficiency, request frequency, and billed cost are distinct quantities.}
\label{fig:token-breakdown}
\end{figure}

The cost study covers 320 complete campaigns: four LLM methods, eight tasks, and ten seeds, each with 100 evaluations. Total tokens are prompt plus completion tokens; prompt tokens are split into provider-reported cache hits and misses. \method{} has the highest total-token usage despite fewer requests. Its repeated context gives a prompt-cache hit rate of 94.0\%, versus 16.4\% for LLAMBO, 27.3\% for Centaur, and 36.2\% for Sara. Rounded component sums can differ slightly from rounded totals. Figure~\ref{fig:token-breakdown} shows the per-task token totals together with the cached-input, uncached-input, and output components, illustrating that total-token efficiency, request frequency, and billed cost are distinct quantities.

Monetary cost requires prices in consistent token units:
\begin{equation}
 C=p_{\mathrm{hit}}T_{\mathrm{hit}}+
 p_{\mathrm{miss}}T_{\mathrm{miss}}+
 p_{\mathrm{out}}T_{\mathrm{out}}.
\end{equation}
Cache hits are still tokens and may incur cost. Neither total tokens nor uncached tokens alone establish which method is cheapest. Counts also exclude numerical computation: fixed HDBO uses no LLM tokens but incurs model-fitting and acquisition-search costs.

\end{document}